\documentclass[acmlarge]{acmart}

\newcommand{\tr}[1]{\multicolumn{1}{r}{#1}} 
\newcommand{\summary}[1]{\vspace{0.3em} \noindent \fbox{\parbox{1\linewidth}{\emph{\textbf{Summary}: #1}}}}

\AtBeginDocument{%
  \providecommand\BibTeX{{%
    \normalfont B\kern-0.5em{\scshape i\kern-0.25em b}\kern-0.8em\TeX}}}

\renewcommand\footnotetextcopyrightpermission[1]{} 
\usepackage{fancyhdr}
\usepackage{caption}
\usepackage{subfigure}
\usepackage{amsmath}
\usepackage{array}
\usepackage{url}
\usepackage{tcolorbox}

\newcommand{\distance}{5pt}
\begin{document}

\title{Demystifying Swarm Learning: A New Paradigm of Blockchain-based Decentralized Federated Learning}

\author{Jialiang Han}
\affiliation{
  \institution{Key Lab of High-Confidence Software Technology, MoE (Peking University), Beijing}
  \country{China}
}
\email{hanjialiang@pku.edu.cn}

\author{Yun Ma}
\authornote{Corresponding author}
\affiliation{%
  \institution{Institute for Artificial Intelligence, Peking University, Beijing}
  \country{China}
}
\email{mayun@pku.edu.cn}

\author{Yudong Han}
\affiliation{
  \institution{Key Lab of High-Confidence Software Technology, MoE (Peking University), Beijing}
  \country{China}
}
\email{hanyd@pku.edu.cn}

\renewcommand{\shortauthors}{Han et al.}


\begin{abstract}
  Federated learning (FL) is an emerging promising privacy-preserving machine learning paradigm and has raised more and more attention from researchers and developers. FL keeps users' private data on devices and exchanges the gradients of local models to cooperatively train a shared Deep Learning (DL) model on central custodians. However, the security and fault tolerance of FL have been increasingly discussed, because its central custodian mechanism or star-shaped architecture can be vulnerable to malicious attacks or software failures. To address these problems, Swarm Learning (SL) introduces a permissioned blockchain to securely onboard members and dynamically elect the leader, which allows performing DL in an extremely decentralized manner. Compared with tremendous attention to SL, there are few empirical studies on SL or blockchain-based decentralized FL, which provide comprehensive knowledge of best practices and precautions of deploying SL in real-world scenarios. Therefore, we conduct the first comprehensive study of SL to date, to fill the knowledge gap between SL deployment and developers, as far as we are concerned. In this paper, we conduct various experiments on 3 public datasets of 5 research questions, present interesting findings, quantitatively analyze the reasons behind these findings, and provide developers and researchers with practical suggestions. The findings have evidenced that SL is supposed to be suitable for most application scenarios, no matter whether the dataset is balanced, polluted, or biased over irrelevant features.
\end{abstract}



\maketitle

\section{Introduction} \label{Section:Introduction}
In many real-world scenarios, such as medical institutions, Internet of Things (IoT) devices, \textit{etc.}, data are distributed in a decentralized manner and the volume of local data is insufficient to train reliable and robust models. It is a common practice to feed local data into a global model and train in a centralized manner, i.e. Centralized Learning (CL), to address the local limitations. Undoubtedly, it raises concerns about data ownership, confidentiality, privacy, security, and monopolies. Recently, the emerging popular Federated Learning (FL)~\cite{DBLP:conf/aistats/McMahanMRHA17} mitigates some of those concerns, where data are kept locally and local confidentiality issues are addressed~\cite{DBLP:conf/ccs/ShokriS15}. Although FL has drawn numerous attention from researchers and developers~\cite{DBLP:conf/www/MaZ00H21, DBLP:conf/www/LiNJZY21, DBLP:conf/www/Liu0C21, DBLP:conf/www/YangWXCBLL21, DBLP:conf/www/ZhangWP21, DBLP:conf/www/Wu0HNWCYZ21}, it uses central custodians to keep model parameters, which could still be attacked to infer users' identities and interests~\cite{DBLP:conf/infocom/WangSZSWQ19, DBLP:conf/sp/NasrSH19, DBLP:journals/jsac/SongWZSWRQ20, DBLP:journals/tifs/ContiMSV16, DBLP:conf/cns/WangYKH15}, even with privacy-preserving Deep Learning (DL) techniques. Besides, the star-shaped architecture of FL damages fault tolerance.

\begin{figure}[htbp]
  \centering
  \includegraphics[width=0.5\textwidth]{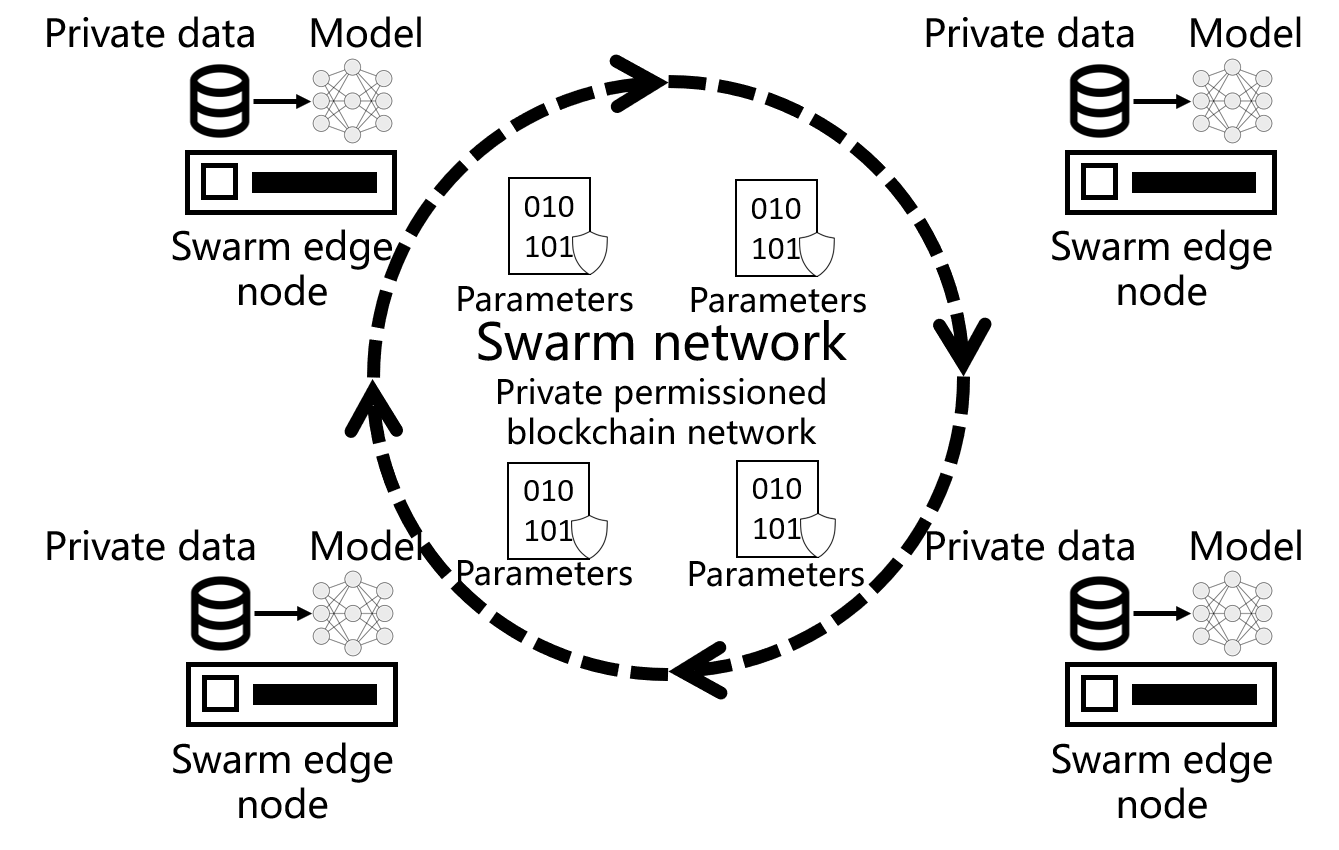}
  \caption{The architecture of Swarm Learning}
  \label{architecture}
  \Description{}
\end{figure}

There are mainly two paths to address the preceding problems raised by FL. One path is on-device training with no raw data or intermediate results uploading~\cite{DBLP:conf/www/Han0ML21, DBLP:journals/imwut/XuQMHL18}, which suffers the over-fitting problem. The other path is decentralized FL~\cite{lalitha2018fully, DBLP:journals/corr/abs-1901-11173, DBLP:conf/blockchain2/RamananN20, DBLP:journals/corr/abs-2101-06905, DBLP:journals/network/LiCLHZY21, warnat2021swarm, warnat2020swarm}, a new FL paradigm to leverage a blockchain to coordinate the model aggregation and update parameters in a decentralized manner. Swarm Learning (SL) is the most representative and state-of-the-art decentralized FL paradigm~\cite{warnat2021swarm}, which combines decentralized hardware infrastructures, distributed machine learning with a permissioned blockchain to securely onboard members, dynamically elect the leader, and merge model parameters. As shown in Figure~\ref{architecture}, SL shares the parameters via the Swarm network and builds the models independently on private data at \textit{Swarm edge nodes}, without the need of a central custodian. With the benefit of blockchain, SL secures data sovereignty, security, and confidentiality. Each participant is well defined and only pre-authorized participants can be onboarded and execute transactions. In the workflow of SL, a new edge node enrolls via a blockchain smart contract, obtains the model, and performs localized training until a user-defined Synchronization Interval (SI). Then, local model parameters are exchanged and merged to update the global model before the next training round. Apart from Swarm edge nodes, there are \textit{Swarm coordinator nodes} responsible for maintaining metadata like the model state, training progress, and licenses, without model parameters.

Despite the increasing attention on the SL paradigm both in industries and academics~\cite{wu2021federated, DBLP:journals/corr/abs-2108-03981, DBLP:journals/corr/abs-2109-12176, oestreich2021privacy, westerlund2021risk}, there lacks comprehensive knowledge of best practices and precautions of deploying SL in real-world scenarios. Although the Swarm Learning Library (SLL) is open-sourced in binary format for non-commercial use\footnote{https://github.com/HewlettPackard/swarm-learning} and has been followed by many researchers and developers, numerous issues about deployment have been raised, such as adaption on different datasets or data distributions, encapsulated features, license assignment, connectivity, heterogeneity in operation systems, hardware infrastructures, DL platforms, and models.

To fill the knowledge gap between SL deployment and developers, in this paper, we use several publicly available datasets and mainstream DL models to measure SL in real-world scenarios, following the methodology in previous measurement studies~\cite{DBLP:conf/imc/XuFMZLQWLYL21, DBLP:conf/www/LinWWL21, DBLP:conf/www/GaoW0LXL21, DBLP:journals/tweb/MaHGZMHL20, DBLP:conf/www/XuLLLLL19, DBLP:conf/www/MaXZTL19}. Specifically, the research questions are listed as follows. On the one hand, how does SL perform in terms of prediction accuracy (\textit{RQ1})? Specifically, considering real-world scenarios, how does SL perform in face of unbalanced datasets (\textit{RQ2}), biased datasets over irrelevant features (\textit{RQ3}), or polluted datasets (\textit{RQ4})? On the other hand, how does SL perform in terms of resource overhead (\textit{RQ1})? Specifically, whether and how much does scaling the nodes affect resource overhead (\textit{RQ5})? Besides, we formulate the SL paradigm, quantitatively analyze possible reasons behind findings, discuss encapsulated features, and provide developers and researchers with practical suggestions. As far as we are concerned, this paper is the \textit{first} comprehensively empirical study focusing on the SL paradigm. Our findings and implications of this paper can be summarized as follows.
\begin{itemize}
    \item Even if the dataset is unbalanced or polluted, developers can trust SL to achieve similar accuracy to CL.
    \item When the dataset is biased over irrelevant features, developers can trust SL to achieve higher fairness than CL. In other words, models trained on different SL nodes provide similar accuracy for the same testing set.
    \item Swarm edge nodes should be deployed on instances with more computational resources and network bandwidth than Swarm coordinator nodes. And some edge nodes consume much more network bandwidth than other edge nodes, perhaps because of the unfairness of the leader election algorithm.
    \item Developers should feel free to increase the number of Swarm coordinator nodes but feel cautious to increase that of Swarm edge nodes, because as the former increases, resource overhead remains basically unchanged, while as the latter increases, resource overhead increases linearly. This means that SL encourages new institution enrollment in the framework, which would hardly consume extra resources.
\end{itemize}


\section{Background} \label{Section:Background}
In this section, we introduce the basic components of SL, how those components interact with each other, and the overall workflow of an SL node.

\subsection{Components} \label{Secion:Background:Components}
There are mainly five components in the SL framework.

\textbf{Swarm Learning (SL) node} runs a user-defined DL algorithm, which iteratively updates the local model. 

\textbf{Swarm Network (SN) node} interacts with each other to maintain global state information about the model and tracks training progress via the Ethereum blockchain platform. Besides, each SL node registers itself with an SN node when initialization and each SN node coordinates the training pipeline of its SL nodes. 
Note that the blockchain records only metadata like the model state and training progress, without model parameters.

\textbf{Swarm Learning Command Interface (SWCI) node} securely connects to SN nodes to view the status, control, and manage the SL framework.

\textbf{SPIFFE SPIRE Server node} guarantees the security of the SL framework. Each SN node or SL node includes a SPIRE\footnote{https://github.com/spiffe/spire} Agent Workload Attestor plugin that communicates with the SPIRE Server nodes to attest their identities and to acquire and manage a SPIFFE\footnote{https://spiffe.io/docs/latest/spiffe-about/overview/} Verifiable Identity Document (SVID).

\textbf{License Server (LS) node} installs and manages the license to run the SL framework.

Note that we refer to SL nodes as Swarm edge nodes, and collectively refer to SN nodes, SWCI nodes, SPIRE Server nodes, and LS nodes as Swarm coordinator nodes in Section~\ref{Section:Introduction} for simplicity and clarity.

\subsection{Component interactions} \label{Secion:Background:Interactions}
\begin{figure}[htbp]
  \centering
  \includegraphics[width=0.5\textwidth]{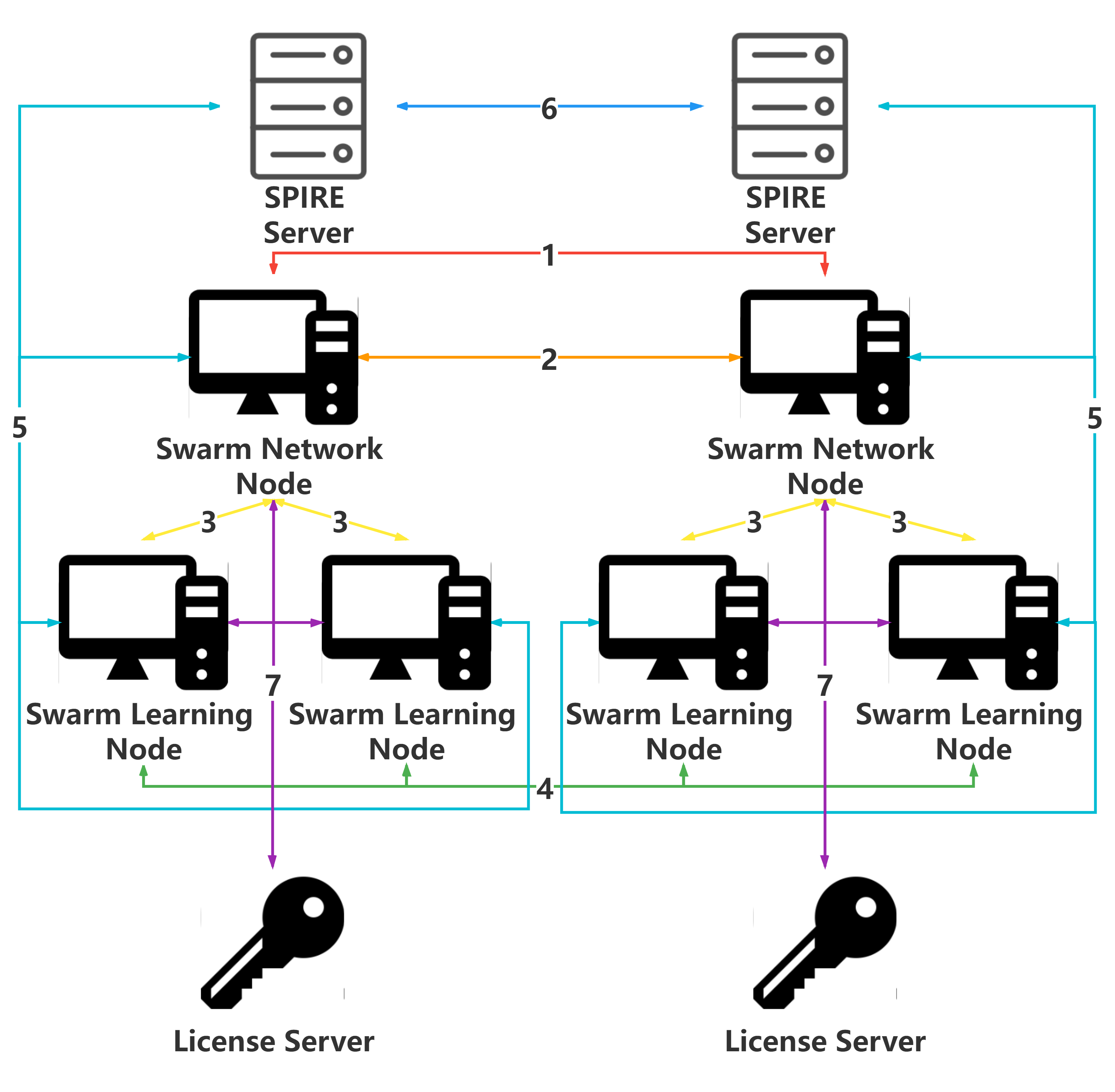}
  \caption{Component interactions of Swarm Learning}
  \label{components}
  \Description{}
\end{figure}

As shown in Figure~\ref{components}, components of SL interact with each other in different ports dedicated for each purpose.

\textbf{Swarm Network Peer-to-Peer Port} is used by SN nodes to share state information about the blockchain with each other, i.e. Line 1 in Figure~\ref{components}.

\textbf{Swarm Network File Server Port} is used by SN nodes to run file servers and share state information about the SL framework with each other, i.e. Line 2 in Figure~\ref{components}.

\textbf{Swarm Network API Port} is used by SN nodes to run REST-based API servers, i.e. Line 3 in Figure~\ref{components}. The API server is used by SL nodes to send and receive state information from the SN node they are registered with. It is also used by SWCI nodes to view and manage the status of the SL framework.

\textbf{Swarm Learning File Server Port} is used by SL nodes to run file servers and share learned parameters of the model with each other, i.e. Line 4 in Figure~\ref{components}.

\textbf{SPIRE Server API Port} is used by SPIRE Server nodes to run gRPC-based API servers for SN nodes and SL nodes to connect to the SPIRE Server and acquire SVIDs, i.e. Line 5 in Figure~\ref{components}.

\textbf{SPIRE Server Federation Port} is used by SPIRE Server nodes to connect each other in the SPIRE federation and send and receive trust bundles, i.e. Line 6 in Figure~\ref{components}.

\textbf{License Server API Port} is used by the LS node to run a REST-based API server and a management interface, i.e. Line 7 in Figure~\ref{components}. The API server is used by SN nodes and SL nodes to connect to the LS node and acquire licenses. The management interface is used by the SL framework administrators to connect to the LS node from browsers and administer licenses.

\subsection{Workflow of a Swarm Learning node} \label{Secion:Background:Workflow}
An SL node initializes in the following pipeline. First, it acquires a license from the LS node. Second, it acquires an SVID from the SPIRE Server node. Third, it registers itself with an SN node. Fourth, it starts a file server and announces to the SN node that it is ready to run the training program. Fifth, it starts running the user-specified model training program.

After initialization, each SL node regularly shares its learned parameters with other SL nodes and merges their parameters. This merging periodicity is defined by a \textit{Synchronization Interval (SI)}, which specifies the number of training batches after which SL nodes merge their learned parameters. 
At the end of each SI, one of the SL nodes is elected as the leader by the blockchain. The leader collects local models from each SL node and merges them into a global model by averaging their learned parameters. After that, each SL node receives this merged model and starts the next SI. 

\section{Methodology} \label{Section:Methodology}
In this section, we introduce how we adapt different models on different public datasets in an SL setting. We set up 3 tasks and 5 research questions to measure the performance of SL in different scenarios. We provide the hyper-parameters in our experimental settings for reproducibility, which we carefully choose through cross validation.

\textbf{Task A: diagnosis of chest X-rays.} We use NIH chest X-ray dataset\footnote{https://nihcc.app.box.com/v/ChestXray-NIHCC} and DenseNet-49 as the classification model. This dataset consists of 112,120 X-ray images with disease labels from 30,805 unique patients~\cite{DBLP:conf/cvpr/WangPLLBS17}. We filter out patients whose ages are above 100 and resize the images to 256*256 pixels.
Note that some samples may have multiple disease labels, therefore, this is a multi-class classification. We use Dense Convolutional Network (DenseNet) as the classifier, which connects layers to each other in a feed-forward fashion. 
DenseNets have several compelling advantages, including alleviating the vanishing-gradient problem, strengthening feature propagation, encouraging feature reuse, and substantially reducing parameters~\cite{DBLP:conf/cvpr/HuangLMW17}. The block sizes of DenseNet-49 are (4, 4, 8, 6). The batch size is 32. The dropout rate of the fully connected layer is 0.15. The initial learning rate is 1e-3, which decays every 40 epochs with the decay ratio $\alpha=0.1$.

\textbf{Task B: object recognition.} We use CIFAR-10 dataset\footnote{https://www.cs.toronto.edu/$\sim$kriz/cifar.html} and DenseNet-BC as the classifier. The CIFAR-10 dataset consists of 60,000 32x32 color images in 10 classes, with 6,000 images per class. The depth of DenseNet-BC is 100 and the growth rate is 12. The batch size is 32. The initial learning rate is 1e-3, which decays at the 80th, 120th, 160th epochs with the decay ratio $\alpha=0.1$.

\textbf{Task C: sentiment analysis.} We use IMDB review dataset\footnote{http://ai.stanford.edu/$\sim$amaas/data/sentiment/} and Attention-based Bidirectional Long Short-Term Memory (Bi-LSTM) as the classifier. The IMDB dataset consists of 50,000 highly polar movie reviews with ratings. Bi-LSTM units have access to both past and future context information and mitigate the vanishing gradient problem to capture long-distance dependencies. And the attention mechanism captures the most important context information in a sentence. The length of the word embedding is 128. The number of Bi-LSTM units is 64. The batch size is 64. The dropout rate of the fully connected layer is 0.05.


\section{Measurement Results} \label{Section:RQs}
In this section, we measure the performance in Section~\ref{Section:RQ1}, the effect of imbalance in Section~\ref{Section:RQ2}, the fairness in Section~\ref{Section:RQ3}, fault tolerance in Section~\ref{Section:RQ4}, and the scalability of SL in Section~\ref{Section:RQ5}.
\subsection{RQ1: Performance} \label{Section:RQ1}
In this research question, we measure the performance of SL, including prediction accuracy, computational overhead, network overhead, and storage overhead.

In each task, we randomly partition the training set into three (four) sub-datasets with an equal number of samples to deploy on three (four) SL nodes. The baseline experimental setting is Centralized Learning (CL) with the whole training set. The testing set of SL and CL is the same for fairness. The detailed statistics are shown in Table~\ref{RQ1.Acc}. The results show that the accuracy of SL is similar or even slightly higher than that of CL. The explanation is that SL introduces ensemble methods that contribute to better generalizability and robustness than CL.
\begin{table}[htbp]
  \caption{Accuracy in RQ1}
  \label{RQ1.Acc}
  \begin{tabular}{m{0.102\columnwidth}m{0.1\columnwidth}m{0.103\columnwidth}m{0.103\columnwidth}m{0.103\columnwidth}m{0.103\columnwidth}m{0.1\columnwidth}}
    \toprule
    &Baseline&Node 1&Node 2&Node 3&Node 4&SL\\
    \midrule
    Task A & \tr{0.8850} & \tr{0.8780} & \tr{0.8768} & \tr{0.8759} & \tr{-} & \tr{0.9090}\\
    Task B & \tr{0.9350} & \tr{0.8922} & \tr{0.8855} & \tr{0.8884} & \tr{-} & \tr{0.9304}\\
    Task C & \tr{0.8940} & \tr{0.8472} & \tr{0.8449} & \tr{0.8543} & \tr{0.8484} & \tr{0.8875}\\
  \bottomrule
\end{tabular}
\end{table}

\begin{figure}[htbp]
\centering
\subfigure[CPU]{
    \begin{minipage}[t]{0.3\textwidth}
        \centering
        \includegraphics[width=\textwidth]{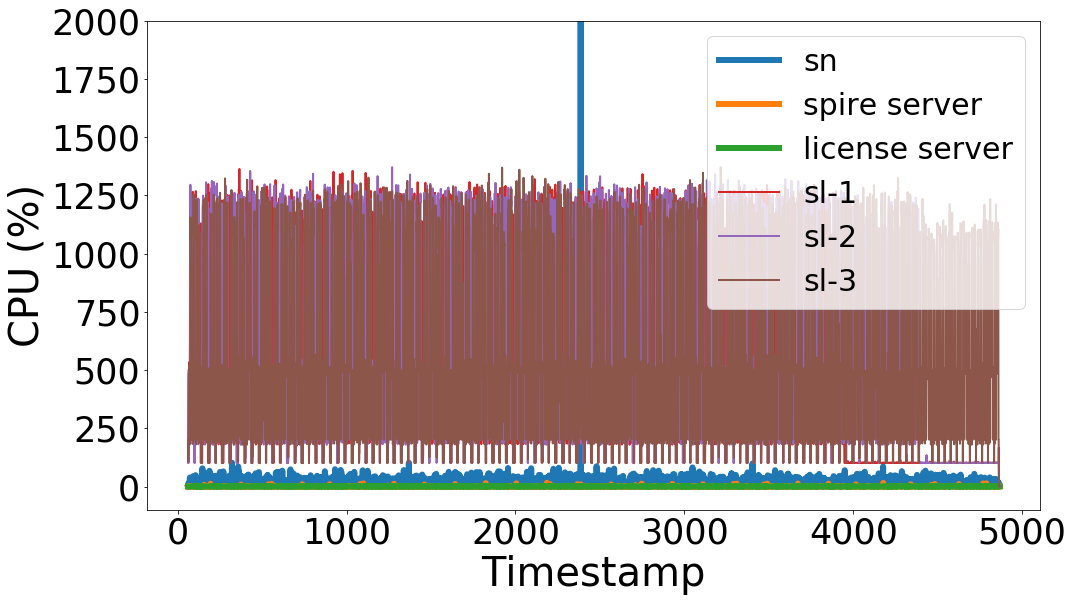}\\
        \vspace{0.02cm}
    \end{minipage}%
}%
\subfigure[Memory]{
    \begin{minipage}[t]{0.3\textwidth}
        \centering
        \includegraphics[width=\textwidth]{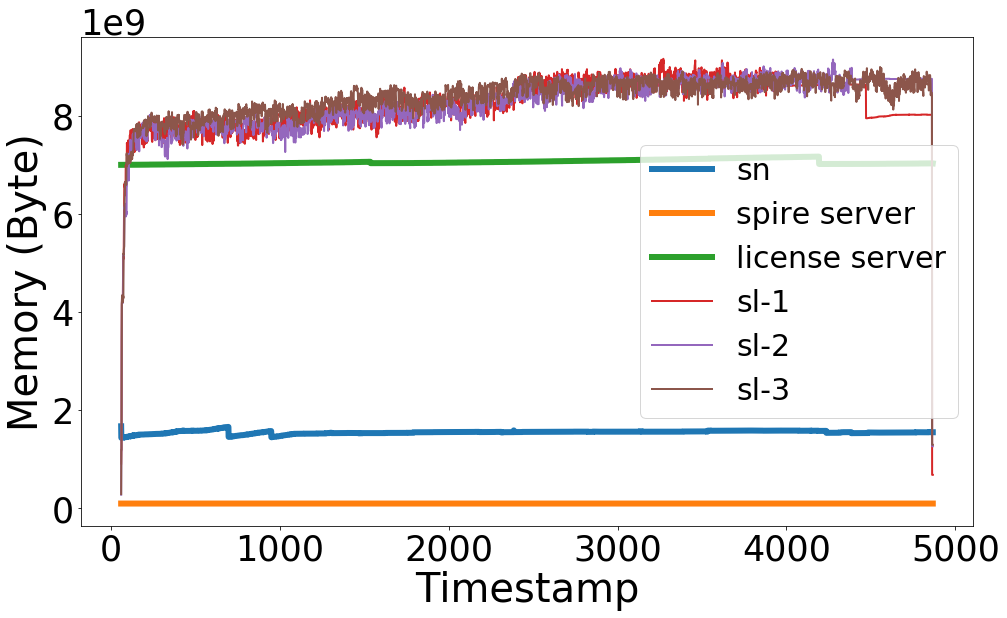}\\
        \vspace{0.02cm}
    \end{minipage}%
}%

\subfigure[Network (in)]{
    \begin{minipage}[t]{0.3\textwidth}
        \centering
        \includegraphics[width=\textwidth]{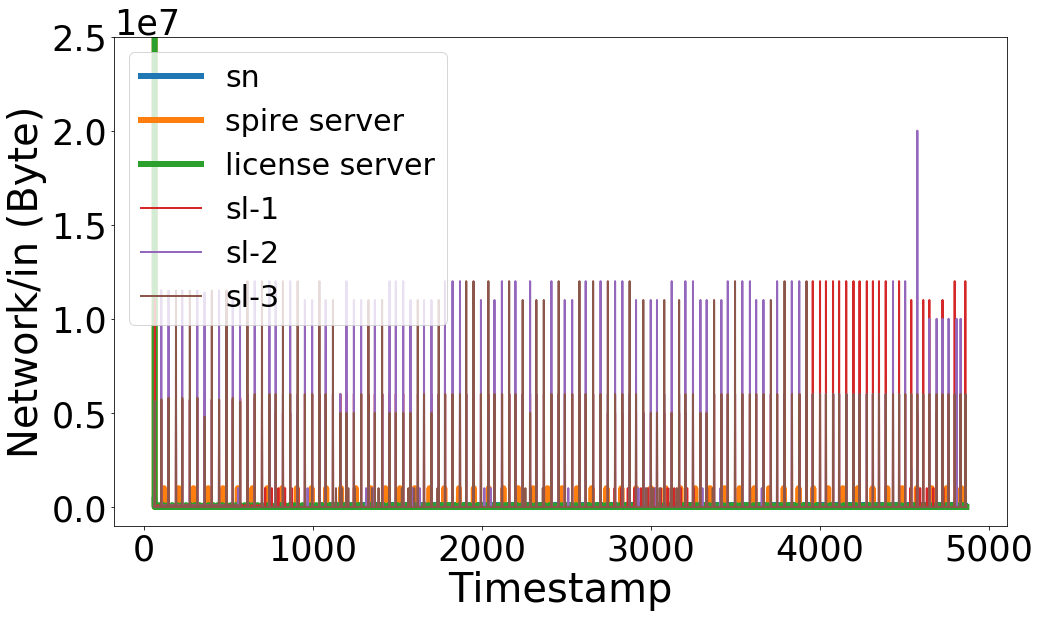}\\
        \vspace{0.02cm}
    \end{minipage}%
}%
\subfigure[Network (out)]{
    \begin{minipage}[t]{0.3\textwidth}
        \centering
        \includegraphics[width=\textwidth]{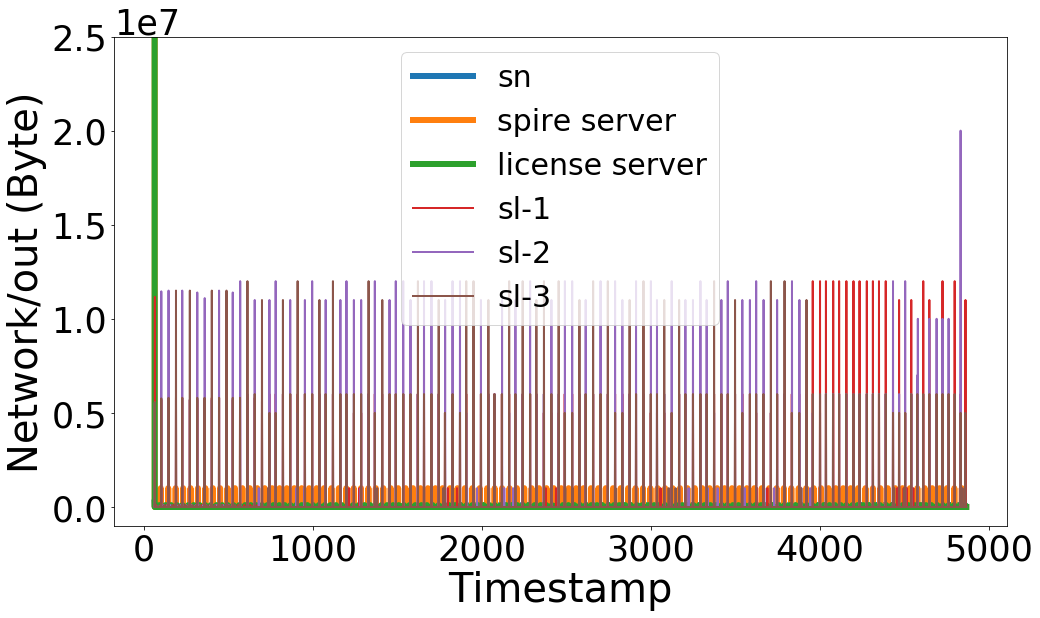}\\
        \vspace{0.02cm}
    \end{minipage}%
}%

\subfigure[Disk (in)]{
    \begin{minipage}[t]{0.3\textwidth}
        \centering
        \includegraphics[width=\textwidth]{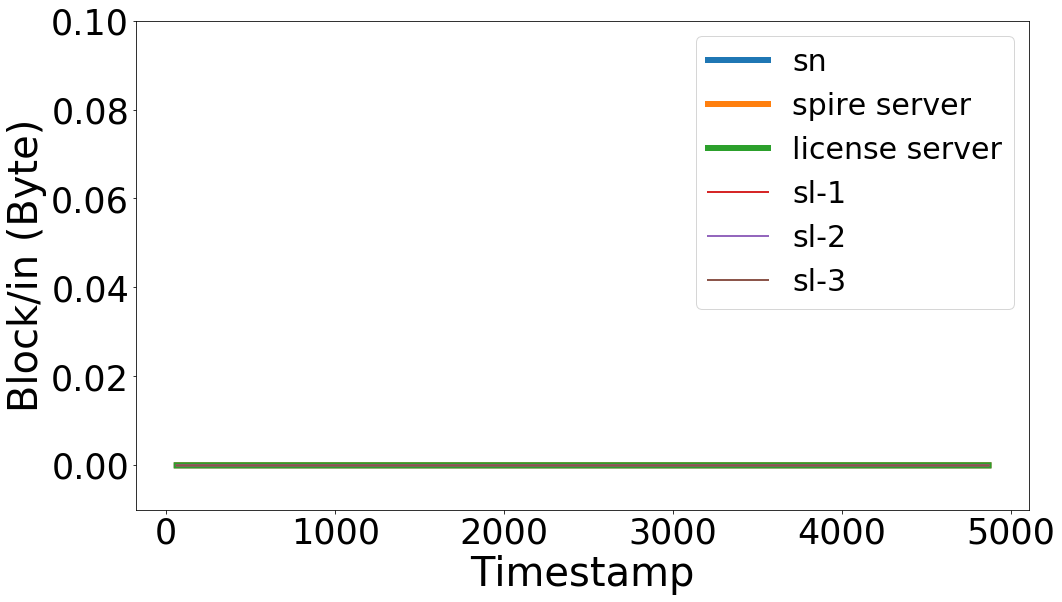}\\
        \vspace{0.02cm}
    \end{minipage}%
}%
\subfigure[Disk (out)]{
    \begin{minipage}[t]{0.3\textwidth}
        \centering
        \includegraphics[width=\textwidth]{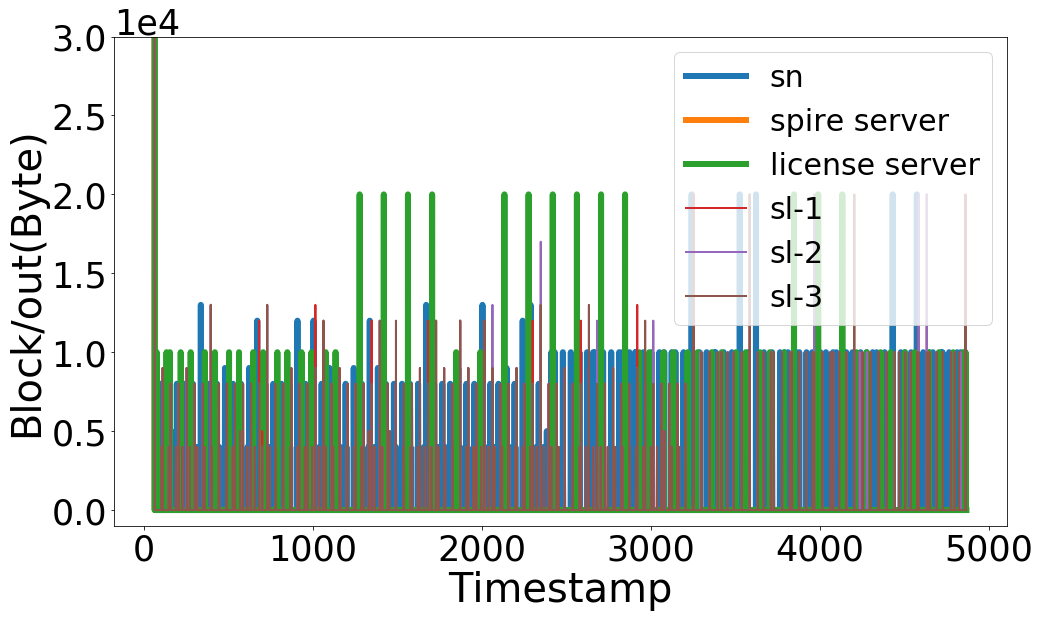}\\
        \vspace{0.02cm}
    \end{minipage}%
}%
\caption{Overhead of Task A in RQ1}
\label{RQ1.TaskA.Overhead}
\end{figure}

The resource overhead of Task A is shown in Figure~\ref{RQ1.TaskA.Overhead}. We notice that SPIRE Server (SS) nodes, License Server (LS) nodes, and Swarm Network (SN) nodes hardly consume CPU resources, while each SL node consumes 500\%-1000\% of single-core CPU resources. From the most to the least memory footprint, each SL node consumes about 8 GB memory, the LS node consumes about 7 GB memory, the SN node consumes about 2 GB memory, while other nodes hardly consume memory resources. As for network overhead, SL nodes take turns consuming 60 to 120 MB per timestamp, i.e. per 5 seconds. The reason behind this take-turn behavior is that the blockchain dynamically elects the leader among members. When a node becomes the current leader, its network overhead would increase significantly because other nodes would send and receive parameters from it. Interestingly, SL-2 consumes much more network bandwidth than other SL nodes. The underlying reason might be the unfairness of the leader election algorithm, which we will discuss in Section~\ref{Section:Implication}. Besides, all kinds of nodes barely consume storage resources. Resource overhead of Task B and Task C is similar to that of Task A. Therefore, we omit these results because of the page limit.

\summary{When the dataset is balanced, the accuracy of SL is similar to that of CL. And SL nodes take turns consuming much more computational resources and network bandwidth than other nodes. Besides, some SL nodes consume much more network bandwidth than other SL nodes.}

\subsection{RQ2: Imbalance} \label{Section:RQ2}
In this research question, we measure the prediction accuracy of SL in face of unbalanced samples on different nodes in Section~\ref{Section:RQ2.1}, and unbalanced labels of samples on each node in Section~\ref{Section:RQ2.2}.

\subsubsection{RQ2.1: Imbalance of Samples on Nodes} \label{Section:RQ2.1}
We randomly partition the training set into three (four) sub-datasets with 1:2:3 (1:2:3:4) samples to deploy on three (four) SL nodes. The other experimental settings are almost the same as those of Section~\ref{Section:RQ1}. The detailed statistics are shown in Table~\ref{RQ2.1}, which show that the accuracy of SL is similar to or would not drop much compared with that of CL.
\begin{table}[htbp]
  \caption{Accuracy in RQ2.1}
  \label{RQ2.1}
  \begin{tabular}{m{0.102\columnwidth}m{0.1\columnwidth}m{0.103\columnwidth}m{0.103\columnwidth}m{0.103\columnwidth}m{0.103\columnwidth}m{0.1\columnwidth}}
    \toprule
    &Baseline&Node 1&Node 2&Node 3&Node 4&SL\\
    \midrule
    Task A & \tr{0.8850} & \tr{0.8750} & \tr{0.8758} & \tr{0.8782} & \tr{-} & \tr{0.8830}\\
    Task B & \tr{0.9350} & \tr{0.8432} & \tr{0.8878} & \tr{0.9086} & \tr{-} & \tr{0.9226}\\
    Task C & \tr{0.8940} & \tr{0.8187} & \tr{0.8437} & \tr{0.8469} & \tr{0.8717} & \tr{0.8943}\\
  \bottomrule
\end{tabular}
\end{table}

\subsubsection{RQ2.2: Imbalance of Labels of Samples}  \label{Section:RQ2.2}
Because Task A is a multi-class classification task, it is difficult to manually truncate this dataset to satisfy the conditions of this research question. 
In Task B, we manually truncate the training set of CIFAR-10 to have 500 to 5000 samples of each class, satisfying the power-law distribution. In Task C, we manually truncate the training set of IMDB to have 12,000 negative samples and 4,000 positive samples. The other experimental settings are almost the same as those of Section~\ref{Section:RQ1}. The detailed statistics are shown in Table~\ref{RQ2.2}, which shows that the accuracy of SL is similar to or even slightly higher than that of CL. The explanation is that the generalizability and robustness improvement of SL benefits accuracy when labels are unbalanced and especially when the dataset is relatively small.
\begin{table}[htbp]
  \caption{Accuracy in RQ2.2}
  \label{RQ2.2}
  \begin{tabular}{m{0.102\columnwidth}m{0.1\columnwidth}m{0.103\columnwidth}m{0.103\columnwidth}m{0.103\columnwidth}m{0.103\columnwidth}m{0.1\columnwidth}}
    \toprule
    &Baseline&Node 1&Node 2&Node 3&Node 4&SL\\
    \midrule
    Task B & \tr{0.8699} & \tr{0.7726} & \tr{0.7606} & \tr{0.7746} & \tr{-} & \tr{0.8559}\\
    Task C & \tr{0.8591} & \tr{0.8146} & \tr{0.8281} & \tr{0.8197} & \tr{0.8061} & \tr{0.8629}\\
  \bottomrule
\end{tabular}
\end{table}

\summary{If SL nodes have different numbers of samples, or the labels of samples are unbalanced, the accuracy of SL would not drop much compared with that of CL.}

\subsection{RQ3: Fairness} \label{Section:RQ3}
In real-world situations, different SL nodes may well have a biased distribution of data over irrelevant features to the prediction results. For example, patients in different medical institutions would have a different distribution of gender, age, ethnicity, \textit{etc}. If those features are irrelevant to the prediction results, models trained on different SL nodes should provide similar accuracy for the same testing set, in consideration of fairness. In this research question, we measure the fairness of SL, i.e. the accuracy of different SL nodes on the testing sets of themselves and others. To achieve this, we randomly partition the dataset on each SL node into a local training set and a local testing set and remain the global testing set mentioned in Section~\ref{Section:RQ1}. Because there are irrelevant features like gender and age in the dataset of Task A, which helps us generate biased sub-datasets, we only test Task A in this research question and generate an NIH-age and NIH-gender sub-dataset groups.

In NIH-age, we partition the training set into three sub-datasets according to the ages of patients. The ages of patients on Node 1 are from 0 to 30, the ages of patients on Node 2 are from 30 to 60, and the ages of patients on Node 3 are from 60 to 100. Note that the number of samples on each node is close to each other. To measure the fairness of SL, we use Localized Learning (LL) as the baseline, where each node performs training on its local training set. 
In both LL and SL settings, each node performs inference on local testing sets of all nodes and the global testing set. We present the Receiver Operation Characteristics Area Under Curve (ROC-AUC) of LL and SL in Table~\ref{RQ3.TaskA.age} to reveal the accuracy gap. The results show that during LL, each node achieves a slightly higher accuracy on its local testing set, which damages the fairness of the model. In contrast, during SL, each node achieves similar accuracy (about 0.71) on Node 1, similar accuracy (about 0.74) on Node 2, and similar accuracy (about 0.74) on Node 3, regardless of whether being its local testing set or not.
\begin{table}[htbp]
  \caption{ROC-AUC of NIH-age in RQ3}
  \label{RQ3.TaskA.age}
  \begin{tabular}{m{0.103\columnwidth}m{0.12\columnwidth}m{0.12\columnwidth}m{0.12\columnwidth}m{0.12\columnwidth}}
    \toprule
    LL/SL&Node 1 Testing&Node 2 Testing&Node 3 Testing&Global Testing\\
    \midrule
    Node 1 & \tr{0.6827/0.7159} & \tr{0.6795/0.7359} & \tr{0.6733/0.7368} & \tr{0.7039/0.7402}\\
    Node 2 & \tr{0.6740/0.7122} & \tr{0.6989/0.7390} & \tr{0.6913/0.7423} & \tr{0.6931/0.7396}\\
    Node 3 & \tr{0.6748/0.7113} & \tr{0.7044/0.7370} & \tr{0.7074/0.7398} & \tr{0.6803/0.7395}\\
  \bottomrule
\end{tabular}
\end{table}

In NIH-gender, we partition the training set into three sub-datasets according to the genders of patients. The female to male ratio of Node 1 is 9:1, the female to male ratio of Node 2 is 5:5, and the female to male ratio of Node 3 is 1:9. Note that the number of samples on each node is close to each other. The other experimental settings are similar to those of Section~\ref{Section:RQ3}. We present the ROC-AUC of Localized Learning (LL) and SL in Table~\ref{RQ3.TaskA.gender}. The results show that during LL, Node 2 achieves a slightly higher accuracy on its own testing set, which damages the fairness of the model. In contrast, during SL, Node 2 and Node 3 achieve similar accuracy (about 0.73) on Node 1, similar accuracy (about 0.74) on Node 2, and similar accuracy (about 0.74) on Node 3, regardless of whether being its own testing set or not. The reason behind the lower accuracy of SL on Node 3 might be that gender is not completely irrelevant to disease labels.
\begin{table}[htbp]
  \caption{ROC-AUC of NIH-gender in RQ3}
  \label{RQ3.TaskA.gender}
  \begin{tabular}{m{0.103\columnwidth}m{0.12\columnwidth}m{0.12\columnwidth}m{0.12\columnwidth}m{0.12\columnwidth}}
    \toprule
    LL/SL&Node 1 Testing&Node 2 Testing&Node 3 Testing&Global Testing\\
    \midrule
    Node 1 & \tr{0.6827/0.7319} & \tr{0.6832/0.7394} & \tr{0.6851/0.7465} & \tr{0.6986/0.7369}\\
    Node 2 & \tr{0.6836/0.7333} & \tr{0.7029/0.7425} & \tr{0.6967/0.7414} & \tr{0.7042/0.7386}\\
    Node 3 & \tr{0.6837/0.7120} & \tr{0.6969/0.7184} & \tr{0.6934/0.7251} & \tr{0.7033/0.7363}\\
  \bottomrule
\end{tabular}
\end{table}

\summary{When the dataset is biased over irrelevant features, models trained on different SL nodes provide similar accuracy for the same testing set. Therefore, the fairness of SL is better than that of CL.}

\subsection{RQ4: Fault Tolerance} \label{Section:RQ4}
In this research question, we measure the fault tolerance of SL in face of low-quality nodes (LQNs), where more than half of samples are incorrectly labeled.

We modify the labels of half of the samples on one node, i.e. LQN, to be incorrect. The baseline is CL with the whole training set, i.e. the union of unpolluted and polluted data. The other experimental settings are almost the same as those of Section~\ref{Section:RQ1}. The detailed statistics are shown in Table~\ref{RQ4}. In Task A and Task B, although the accuracy of LQN is much lower than other nodes, the accuracy of SL is similar to or even slightly higher than that of CL. The reason behind this is the mechanism of training separately on unpolluted and polluted data and then merging periodically, and that localized training on unpolluted data would mitigate the adverse effect of polluted data on accuracy. However, in Task C, the accuracy of SL is lower than that of CL. The reason behind this is that the IMDB dataset is relatively small and the model on each node is easy to converge, however, the SL model cannot converge eventually, because the models trained on unpolluted and polluted data are quite different. Even so, SL mitigates the adverse effect of polluted data to some extent because the accuracy of the LQN is even much lower than that of SL.
\begin{table}[htbp]
  \caption{Accuracy in RQ4}
  \label{RQ4}
  \begin{tabular}{m{0.102\columnwidth}m{0.1\columnwidth}m{0.103\columnwidth}m{0.103\columnwidth}m{0.103\columnwidth}m{0.103\columnwidth}m{0.1\columnwidth}}
    \toprule
    Metric&Baseline&Node 1&Node 2&Node 3&LQN&SL\\
    \midrule
    Task A & \tr{0.8841} & \tr{0.8795} & \tr{0.8805} & \tr{-} & \tr{0.8796} & \tr{0.8840}\\
    Task B & \tr{0.8673} & \tr{0.8933} & \tr{0.8880} & \tr{-} & \tr{0.4480} & \tr{0.8897}\\
    Task C & \tr{0.8646} & \tr{0.8322} & \tr{0.8500} & \tr{0.8380} & \tr{0.5002} & \tr{0.7955}\\
  \bottomrule
\end{tabular}
\end{table}

\summary{If there exist low-quality nodes, where more than half of samples are incorrectly labeled, in most scenarios, the accuracy of SL would be similar to that of CL.}

\subsection{RQ5: Scalability} \label{Section:RQ5}
In this research question, we measure the scalability of SL, i.e. computational overhead, network overhead, and storage overhead as SN nodes scale in Section~\ref{Section:RQ5.1} or as SL nodes scale in Section~\ref{Section:RQ5.2}. Note that the number of SN nodes here refers to the total number of SL nodes attached to all SN nodes. To measure computational overhead, we measure the average usage percentage of CPU, the average memory footprint. To measure network overhead, we measure the amount of data which have been sent or received over the network interface. To measure storage overhead, we measure the amount of data which have been read from or written to block devices. Note that as for network and storage overhead, we measure the total amount of data in or out, instead of the average, because the convergence time of training on different numbers of SL nodes varies significantly when the dataset is fixed, which makes measuring the total overhead more reasonable than the average. We measure the above resource overhead on SWCI nodes, SPIRE Server (SS) nodes, License Server (LS) nodes, SN nodes, and SL nodes mentioned in Figure~\ref{components}. Due to the page limit, we only show the results of Task B. Because more SL nodes consume more GPU memory, we choose EfficientNetB2~\cite{DBLP:conf/icml/TanL19} to save GPU memory instead of DenseNet-BC in previous research questions for Task B.

\subsubsection{RQ5.1: Scaling SN nodes} \label{Section:RQ5.1}
We fix the total number of SL nodes to 4 and vary the number of SN nodes from 1 to 4.

\textbf{Scenario 1.} When the number of SN nodes is 1, 4 SL nodes are linked to each SN node. The detailed statistics are shown in Table~\ref{RQ5.1.1SN}. The findings are similar to those in Section~\ref{Section:RQ1}, only that the absolute values are different.
\begin{table}[htbp]
  \caption{Overhead of Task B (SN = 1, SL = 4)}
  \label{RQ5.1.1SN}
  \begin{tabular}{m{0.1\columnwidth}m{0.1\columnwidth}m{0.1\columnwidth}m{0.1\columnwidth}m{0.1\columnwidth}m{0.1\columnwidth}m{0.1\columnwidth}}
    \toprule
    Node&CPU (\%)&Mem (MB)&Net/in (MB)&Net/out (MB)&Block/in (MB)&Block/out (MB)\\
    \midrule
    SWCI & \tr{0.26} & \tr{91.29} & \tr{10.4} & \tr{4.2} & \tr{3.08} & \tr{0.02}\\
    SS & \tr{0.68} & \tr{143.19} & \tr{23.7} & \tr{58.3} & \tr{89.60} & \tr{0.02}\\
    LS & \tr{0.36} & \tr{2392.41} & \tr{6.5} & \tr{5.1} & \tr{120.00} & \tr{0.73}\\
    SN-0 & \tr{11.25} & \tr{1386.77} & \tr{28.9} & \tr{17.7} & \tr{0.11} & \tr{1.76}\\
    SL-0-0 & \tr{205.56} & \tr{2920.06} & \tr{8820.0} & \tr{8820.0} & \tr{0.11} & \tr{1.56}\\
    SL-0-1 & \tr{205.20} & \tr{2978.02} & \tr{10700.0} & \tr{10700.0} & \tr{0.05} & \tr{1.56}\\
    SL-0-2 & \tr{206.58} & \tr{3034.54} & \tr{17800.0} & \tr{17800.0} & \tr{0.00} & \tr{1.56}\\
    SL-0-3 & \tr{206.38} & \tr{3013.60} & \tr{15200.0} & \tr{15200.0} & \tr{0.01} & \tr{1.56}\\
  \bottomrule
\end{tabular}
\end{table}

\textbf{Scenario 2.} When the number of SN nodes is 2, 2 SL nodes are linked to each SN node. The detailed statistics are shown in Table~\ref{RQ5.1.2SN}. In terms of computational overhead, we notice similar findings to Scenario 1, except that two SN nodes consume significantly different CPU and memory resources from each other. In terms of network overhead, the SWCI node, the SS node, the LS node consume about 2 times network resources as many as Scenario 1. Each SN node and each SL node consume similar network resources to Scenario 1, except SL-0-1, perhaps because SL-0-1 is elected as the leader more frequently than others. In terms of storage overhead, each node consumes similar storage resources to Scenario 1.
\begin{table}[htbp]
  \caption{Overhead of Task B (SN = 2, SL = 4)}
  \label{RQ5.1.2SN}
  \begin{tabular}{m{0.1\columnwidth}m{0.1\columnwidth}m{0.1\columnwidth}m{0.1\columnwidth}m{0.1\columnwidth}m{0.1\columnwidth}m{0.1\columnwidth}}
    \toprule
    Node&CPU (\%)&Mem (MB)&Net/in (MB)&Net/out (MB)&Block/in (MB)&Block/out (MB)\\
    \midrule
    SWCI & \tr{0.21} & \tr{93.89} & \tr{18.2} & \tr{8.11} & \tr{3.08} & \tr{0.05}\\
    SS & \tr{0.78} & \tr{146.34} & \tr{49.2} & \tr{109.0} & \tr{90.30} & \tr{0.02}\\
    LS & \tr{0.38} & \tr{3102.69} & \tr{13.7} & \tr{10.7} & \tr{120.00} & \tr{1.10}\\
    SN-0 & \tr{10.79} & \tr{1295.24} & \tr{28.3} & \tr{29.2} & \tr{0.14} & \tr{1.71}\\
    SN-1 & \tr{2.73} & \tr{658.53} & \tr{35.0} & \tr{20.7} & \tr{0.00} & \tr{0.59}\\
    SL-0-0 & \tr{201.98} & \tr{3029.52} & \tr{9120.0} & \tr{9120.0} & \tr{0.02} & \tr{1.56}\\
    SL-0-1 & \tr{203.47} & \tr{3035.19} & \tr{25100.0} & \tr{25100.0} & \tr{0.00} & \tr{1.56}\\
    SL-1-2 & \tr{201.35} & \tr{2956.23} & \tr{9640.0} & \tr{9640.0} & \tr{0.24} & \tr{1.56}\\
    SL-1-3 & \tr{206.07} & \tr{2934.75} & \tr{8770.0} & \tr{8770.0} & \tr{0.00} & \tr{1.56}\\
  \bottomrule
\end{tabular}
\end{table}

\textbf{Scenario 3.} When the number of SN nodes is 4, only 1 SL node is linked to each SN node. The detailed statistics are shown in Table~\ref{RQ5.1.4SN}. In terms of computational overhead, we notice similar findings to Scenario 1, except that SN-0 consumes significantly more CPU and memory resources than others. In terms of network overhead, the SWCI node, the SS node, the LS node consume about 4 times network resources as many as Scenario 1. Each SN node and each SL node consume similar network resources to Scenario 1, except SL-1-1. And the reason behind this is the same as that when the number of SN nodes grows from 1 to 2. In terms of storage overhead, each node consumes similar storage resources to Scenario 1.
\begin{table}[htbp]
  \caption{Overhead of Task B (SN = 4, SL = 4)}
  \label{RQ5.1.4SN}
  \begin{tabular}{m{0.1\columnwidth}m{0.1\columnwidth}m{0.1\columnwidth}m{0.1\columnwidth}m{0.1\columnwidth}m{0.1\columnwidth}m{0.1\columnwidth}}
    \toprule
    Node&CPU (\%)&Mem (MB)&Net/in (MB)&Net/out (MB)&Block/in (MB)&Block/out (MB)\\
    \midrule
    SWCI & \tr{0.46} & \tr{94.89} & \tr{39.3} & \tr{18.7} & \tr{3.08} & \tr{0.05}\\
    SS & \tr{0.63} & \tr{149.55} & \tr{103.0} & \tr{216.0} & \tr{91.00} & \tr{0.25}\\
    LS & \tr{0.32} & \tr{4025.16} & \tr{28.0} & \tr{21.9} & \tr{120.00} & \tr{1.74}\\
    SN-0 & \tr{10.58} & \tr{1442.41} & \tr{45.0} & \tr{64.3} & \tr{0.25} & \tr{0.59}\\
    SN-1 & \tr{2.60} & \tr{752.57} & \tr{34.7} & \tr{20.4} & \tr{0.13} & \tr{1.53}\\
    SN-2 & \tr{2.43} & \tr{659.89} & \tr{32.2} & \tr{18.6} & \tr{0.11} & \tr{0.77}\\
    SN-3 & \tr{2.22} & \tr{635.66} & \tr{30.5} & \tr{17.7} & \tr{0.00} & \tr{0.56}\\
    SL-0-0 & \tr{204.77} & \tr{2991.51} & \tr{9000.0} & \tr{9010.00} & \tr{0.00} & \tr{1.56}\\
    SL-1-1 & \tr{202.29} & \tr{2856.76} & \tr{22300.0} & \tr{22300.00} & \tr{0.00} & \tr{1.58}\\
    SL-2-2 & \tr{203.00} & \tr{2821.84} & \tr{12000.0} & \tr{12000.00} & \tr{0.01} & \tr{1.55}\\
    SL-3-3 & \tr{203.46} & \tr{2939.63} & \tr{9230.0} & \tr{9240.00} & \tr{0.00} & \tr{1.55}\\
  \bottomrule
\end{tabular}
\end{table}

In summary, as the number of SN nodes increases, resource overhead remains basically unchanged, because SL nodes account for the most resource overhead.

\subsubsection{RQ5.2: Scaling SL nodes} \label{Section:RQ5.2}
We fix the number of SN nodes to 2 and vary the number of SL nodes from 2 to 8.

\textbf{Scenario 4.} When the number of SL nodes is 2, only 1 SL node is linked to each SN node. The detailed statistics are shown in Table~\ref{RQ5.2.2SL}. 
The findings are similar to those in Section~\ref{Section:RQ1}, only that the absolute values are different.
\begin{table}[htbp]
  \caption{Overhead of Task B (SN = 2, SL = 2)}
  \label{RQ5.2.2SL}
  \begin{tabular}{m{0.1\columnwidth}m{0.1\columnwidth}m{0.1\columnwidth}m{0.1\columnwidth}m{0.1\columnwidth}m{0.1\columnwidth}m{0.1\columnwidth}}
    \toprule
    Node&CPU (\%)&Mem (MB)&Net/in (MB)&Net/out (MB)&Block/in (MB)&Block/out (MB)\\
    \midrule
    SWCI & \tr{0.20} & \tr{96.88} & \tr{78.7} & \tr{38.4} & \tr{3.08} & \tr{0.07}\\
    SS & \tr{0.49} & \tr{153.28} & \tr{208.0} & \tr{427.0} & \tr{91.20} & \tr{0.25}\\
    LS & \tr{0.21} & \tr{4341.82} & \tr{56.9} & \tr{46.5} & \tr{127.00} & \tr{3.14}\\
    SN-0 & \tr{9.70} & \tr{1528.33} & \tr{27.8} & \tr{31.8} & \tr{0.00} & \tr{1.86}\\
    SN-1 & \tr{2.02} & \tr{662.85} & \tr{38.2} & \tr{22.2} & \tr{0.00} & \tr{0.63}\\
    SL-0-0 & \tr{226.36} & \tr{3244.31} & \tr{8780.0} & \tr{8760.00} & \tr{0.00} & \tr{1.66}\\
    SL-1-1 & \tr{226.61} & \tr{3241.23} & \tr{8760.0} & \tr{8780.00} & \tr{0.01} & \tr{1.66}\\
  \bottomrule
\end{tabular}
\end{table}

When the number of SL nodes is 4, 2 SL nodes are linked to each SN node. This experimental setting is the same as Scenario 2 in Section~\ref{Section:RQ5.1}. Therefore, the detailed statistics are shown in Table~\ref{RQ5.1.2SN}. In terms of computational overhead, the SWCI node, the SS node, the LS node, and SN nodes consume similar CPU and memory resources to Scenario 4. The computational overhead of each SL node decreases slightly. In terms of network overhead, the SWCI node, the SS node, the LS node, each SN node, and most SL nodes consume similar network bandwidth to Scenario 4. However, SL-0-1 consumes about 2.5 times network bandwidth as much as Scenario 4, perhaps because SL-0-1 is elected as the leader more frequently than others. Besides, each node consumes similar storage resources to Scenario 4.

\textbf{Scenario 5.} When the number of SL nodes is 8, 4 SL nodes are linked to each SN node. The detailed statistics are shown in Table~\ref{RQ5.2.8SL}. In terms of computational overhead, the SWCI node, the SS node, the LS node, and SN nodes consume similar CPU and memory resources to Scenario 4 in Section~\ref{Section:RQ5.2}. The computational overhead of each SL node decreases slightly. In terms of network overhead, the SWCI node, the SS node, the LS node, each SN node, and most SL nodes consume similar network resources to Scenario 4, except that SL-0-1 consumes about 1.5 times network resources as many as Scenario 4, and SL-0-2 consumes about 5 times network resources as many as Scenario 4. And the reason behind this is the same as that when the number of SL nodes grows from 2 to 4. Besides, each node consumes similar storage resources to Scenario 4.
\begin{table}[htbp]
  \caption{Overhead of Task B (SN = 2, SL = 8)}
  \label{RQ5.2.8SL}
  \begin{tabular}{m{0.1\columnwidth}m{0.1\columnwidth}m{0.1\columnwidth}m{0.1\columnwidth}m{0.1\columnwidth}m{0.1\columnwidth}m{0.1\columnwidth}}
    \toprule
    Node&CPU (\%)&Mem (MB)&Net/in (MB)&Net/out (MB)&Block/in (MB)&Block/out (MB)\\
    \midrule
    SWCI & \tr{0.29} & \tr{96.52} & \tr{56.5} & \tr{27.3} & \tr{3.08} & \tr{0.07}\\
    SS & \tr{0.78} & \tr{153.39} & \tr{165.0} & \tr{340.0} & \tr{91.10} & \tr{0.25}\\
    LS & \tr{0.40} & \tr{4228.73} & \tr{46.1} & \tr{38.1} & \tr{127.00} & \tr{2.69}\\
    SN-0 & \tr{11.68} & \tr{1626.26} & \tr{44.7} & \tr{41.0} & \tr{0.00} & \tr{1.74}\\
    SN-1 & \tr{3.89} & \tr{675.86} & \tr{50.9} & \tr{31.5} & \tr{0.00} & \tr{0.58}\\
    SL-0-0 & \tr{176.14} & \tr{2896.07} & \tr{10500.0} & \tr{10500.0} & \tr{0.14} & \tr{1.57}\\
    SL-0-1 & \tr{176.64} & \tr{2926.51} & \tr{14900.0} & \tr{14900.0} & \tr{0.02} & \tr{1.56}\\
    SL-0-2 & \tr{182.21} & \tr{2807.66} & \tr{51100.0} & \tr{51100.0} & \tr{0.13} & \tr{1.56}\\
    SL-0-3 & \tr{175.33} & \tr{2876.95} & \tr{9100.0} & \tr{9110.0} & \tr{0.00} & \tr{1.56}\\
    SL-1-4 & \tr{176.72} & \tr{2870.63} & \tr{9980.0} & \tr{9980.0} & \tr{0.00} & \tr{1.57}\\
    SL-1-5 & \tr{175.20} & \tr{2834.57} & \tr{8760.0} & \tr{8760.0} & \tr{0.00} & \tr{1.56}\\
    SL-1-6 & \tr{175.48} & \tr{2899.40} & \tr{9450.0} & \tr{9450.0} & \tr{0.02} & \tr{1.57}\\
    SL-1-7 & \tr{175.94} & \tr{2670.77} & \tr{8760.0} & \tr{8760.0} & \tr{0.00} & \tr{1.57}\\
  \bottomrule
\end{tabular}
\end{table}

In summary, as the number of SL nodes increases, computational overhead increases linearly (but slower), because the sub-dataset on each SL node grows smaller. And network overhead increases linearly (but faster) because dividing the dataset into more SL nodes introduces extra difficulties in convergence and some SL nodes are elected as the leader more frequently than others. The storage overhead of each node remains unchanged.

\summary{As the number of SN nodes increases, resource overhead remains basically unchanged. As the number of SL nodes increases, computational overhead and network overhead increase linearly, while storage overhead remains unchanged.}

\section{Formulation and Quantitative Analysis} \label{Section:Formulation}
In Section~\ref{Section:RQs}, we mostly analyze experimental results qualitatively. In this section, we formulate the SL optimization problem and analyze the above results quantitatively.

The objective of SL is to minimize the weighted average loss $f(w)=\sum_{k=1}^Kp_kF_k(w)=\sum_{k=1}^K\frac{n_k}{n}F_k(w)$, where $w$ is model parameters, and data are distributed on $K$ peers, with $\mathcal{P}_k$ the set of samples on peer $k$, with $n_k=|\mathcal{P}_k|$. The local objective of a SL peer is to minimize the average loss $F_k(w)=\frac{1}{n_k}\sum_{i\in\mathcal{P}_k}f_i(w)$, where $f_i(w)=\ell(x_i,y_i;w)$. During each Synchronization Interval (SI), each peer locally takes several batches of Stochastic Gradient Descents (SGDs) on the current model using its local data, and then the leader takes a weighted average of the resulting parameters and assigns it to each peer. This model aggregation process is similar to \texttt{FedAvg} in FL~\cite{DBLP:conf/aistats/McMahanMRHA17}, except that the leader is elected dynamically through a blockchain instead of using a fixed central custodian. Therefore, the convergence analysis of FL~\cite{DBLP:conf/iclr/LiHYWZ20} can be applied to SL, where \texttt{FedAvg} has $\mathcal{O}(\frac{1}{T})$ convergence rate for strongly convex and smooth problems, where $T$ is the number of SGDs. Besides, there is a trade-off between communication efficiency and convergence rate, even with non-iid data (i.e. data are not identically distributed) or partial device participation. Theoretically, this explains the findings that the convergence accuracy of SL is similar to that of Centralized Learning (CL) in most scenarios and reminds developers to choose an optimal SI. The reason why the accuracy of SL is sometimes higher than that of CL is that ensemble methods benefit the robustness and reach a better local optimal point for non-convex objectives~\cite{DBLP:conf/nips/ChenLSS17}. Besides, the reason behind the fairness of SL is that SL models converge to optimal points of the global training set and Localized Learning (LL) models converge to those of the local training set.

As for scalability, because SN nodes are not responsible for model training, resource overhead remains basically unchanged as SN nodes scale. Besides, the reason behind the changes of computational overhead and storage overhead as SL nodes scale is apparent and explained in Section~\ref{Section:RQ5.2}, while the reason behind the changes of network overhead is interesting. As mentioned in the convergence analysis of FL~\cite{DBLP:conf/iclr/LiHYWZ20}, the number of communications is formulated as
\begin{equation}
    \frac{T}{E}=\mathcal{O}\left[\frac{1}{\epsilon}\left(\left(1+\frac{1}{K}\right)EG^2+\frac{\sum_{k=1}^Kp_k^2\sigma_k^2+\Gamma+G^2}{E}+G^2\right)\right],
\end{equation}
where $E$ is the number of local iterations performed in a peer between synchronizations, $\epsilon$ is a fixed precision, $G$ and $\sigma_k$ are problem-related boundaries, $\Gamma$ quantifies the degree of non-iid, and $p_k=\frac{1}{K}$ when the distribution of data is balanced. Therefore, the convergence network overhead of SL is formulated as
\begin{equation} \label{equation:NO}
\begin{aligned}
    \mathcal{NO}=\mathcal{M}K\frac{T}{E}&=\mathcal{O}\left[\frac{\mathcal{M}}{\epsilon}\left(\left(K+1\right)EG^2+\frac{\sigma_k^2+K\Gamma+KG^2}{E}+KG^2\right)\right]\\
    &=\mathcal{O}\left[\frac{\mathcal{M}}{\epsilon}\left(\left((E+1)G^2+\frac{\Gamma+G^2}{E}\right)K+EG^2+\frac{\sigma_k^2}{E}\right)\right],\\
\end{aligned}
\end{equation}
where $\mathcal{M}$ is the size of the model. Equation~\ref{equation:NO} explains the linear increase and the increase slope of network overhead as the number of SL nodes $K$ scales. Besides, Equation~\ref{equation:NO} can be easily extended to the scenario where the distribution of data is unbalanced.

\section{Implication} \label{Section:Implication}
According to findings in Section~\ref{Section:RQs}, we provide suggestions to developers deploying SL on real-world applications on what is recommended to do and what is not recommended to do. Besides, we provide suggestions to researchers on what to improve in SL.

\subsection{What to Do?}
No matter whether the dataset is balanced on SL nodes, developers can trust SL to achieve similar accuracy to Centralized Learning (CL), according to Section~\ref{Section:RQ1} and Section~\ref{Section:RQ2}. No matter whether the dataset is biased over irrelevant features to the prediction results, developers can trust SL to achieve higher fairness than CL, according to Section~\ref{Section:RQ3}.

If developers are uncertain whether there exist low-quality nodes (LQNs) with polluted data, they should pay attention to the size of the dataset and the convergence of SL. If the dataset is sufficient enough, SL is more likely to converge, and developers can trust SL to achieve similar accuracy to CL, according to Section~\ref{Section:RQ4}.

As for deployment, according to Section~\ref{Section:RQ1} and Section~\ref{Section:RQ5}, SL nodes should be deployed on instances with sufficient CPU resources, memory, and network bandwidth. SN nodes and License Server (LS) nodes should be deployed on instances with sufficient memory. Other nodes have few special requirements of all resources, which indicates that developers can deploy these nodes on economical instances to save costs.

Developers should feel free to increase the number of SN nodes, with few worries about consuming extra resources, according to Section~\ref{Section:RQ5.1}. This implication is meaningful in real-world applications because each SN node represents an institution with new data and resources. For example, in a scenario that medical institutions around the world are designing an effective vaccine of COVID-19 based on SL, a newly enrolled medical institution can contribute to the research and development (R \& D) process without requiring extra resources from other institutions, which encourages more institutions to participate and discourages existing institutions from quitting.

\subsection{What Not to Do?}
When there exist LQNs with polluted data and the dataset is relatively small, SL is more likely to not converge and fail to achieve similar accuracy to CL, according to Section~\ref{Section:RQ4}. Therefore, developers are not recommended to perform SL in this scenario.

Peers in the SL framework should feel cautious to increase the number of SL nodes. According to Section~\ref{Section:RQ5.2}, as the number of SL nodes increases, network overhead increases linearly, but faster. Blindly adding SL nodes in one SN node would indeed save training time, but significantly increase network overhead. Therefore, it is not recommended to blindly add SL nodes in one SN node.

\subsection{Where to Improve?}
According to Section~\ref{Section:RQ1} and Section~\ref{Section:RQ5}, some SL nodes consume much more network bandwidth than others. This finding leads to two disadvantages. First, peers with more overhead would complain about that unfairness and tend to quit from the SL framework. Second, if hackers detect which SL node sends or receives much more data than others, they would assume that this node is elected as the leader more frequently, and this would reduce the difficulty from attacking all SL nodes to attacking only one SL node with the most overhead, to infer users' identities and interests. As mentioned in Section~\ref{Section:Discussion}, the Leader Election Algorithm (LEA) is not open-sourced. Therefore, we assume that the underlying reason might be the unfairness of LEA.

This unfairness means that the probability of different SL nodes being elected as the leader is not equivalent or proportional to computing power. Under this assumption, LEA could be Proof of Stake (PoS)~\cite{DBLP:conf/bigdata/ZhengXDCW17} or its variants, where the leader election possibility is proportional to stakes or account balance of nodes. Therefore, it leads to severe unfairness when stakes on SL nodes are significantly different for some reasons. To mitigate the unfairness of PoS, we can change the LEA into Proof of Work (PoW)~\cite{nakamoto2008bitcoin} or its variants. In our PoW setting, each SL node elects itself as the leader and frequently changes its block header to get different hash values. When one SL node reaches the consensus that the calculated hash value is equal to or smaller than a certain given value, it would broadcast to other nodes that it is the current leader. The difficulty of reaching consensus can be controlled by targeting different amounts of hash value bits to satisfy requirements. Theoretically, PoW ensures the leader election probability to be proportional to computing power. And when SL nodes share similar computing power, this refinement for LEA ensures this probability to be equivalent.

\begin{figure}[htbp]
\centering
\subfigure[PoS-based LEA with different stakes among nodes]{
    \begin{minipage}[t]{0.3\textwidth}
        \centering
        \includegraphics[width=\textwidth]{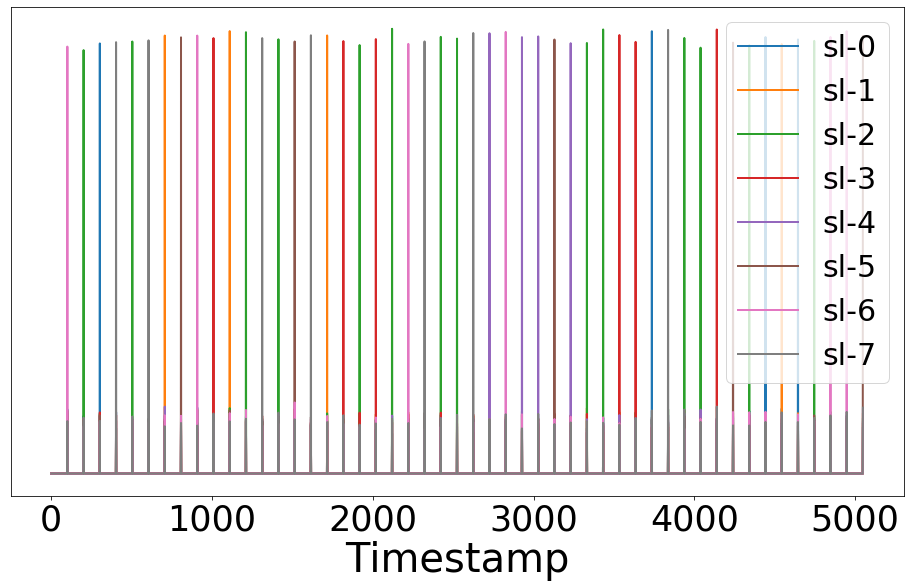}\\
        \vspace{0.02cm}
    \end{minipage}%
}%
\subfigure[PoW-based LEA]{
    \begin{minipage}[t]{0.3\textwidth}
        \centering
        \includegraphics[width=\textwidth]{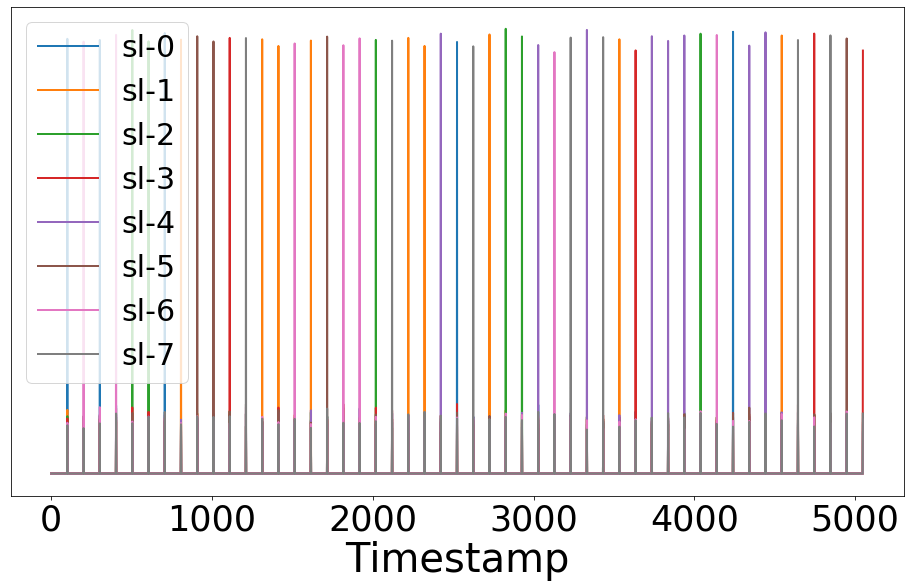}\\
        \vspace{0.02cm}
    \end{minipage}%
}%
\caption{Network Overhead of Different LEAs}
\label{Implication.Election.Overhead}
\end{figure}

We also conduct simulation experiments to measure the network overhead of the PoS-based LEA with different stakes among nodes and the PoW-based LEA. In the PoS-based LEA, according to the results in Section~\ref{Section:RQ5.2} and the above assumption, we assign stakes proportional to the network overhead of SL nodes. As shown in Figure~\ref{Implication.Election.Overhead}, in the PoS-based LEA, SL-1 and SL-2 are elected as leaders more frequently than others and consume much more network bandwidth than others, which is consistent with the observations in Section~\ref{Section:RQ1} and Section~\ref{Section:RQ5}. In contrast, in the PoW-based LEA, the frequency of being elected as the leader and network overhead of SL nodes are close to each other. Therefore, the results of simulation experiments would verify our assumption and refinements. In future, we will try to cooperate with Hewlett Packard Enterprise (HPE) to improve the fairness of the LEA in SL.

\section{Threats to Validity} \label{Section:Discussion}
In this section, we discuss some threats to our methodology and experimental results of research questions.

\textbf{Black-box testing.} As we mention in Section~\ref{Section:RelatedWork}, although Hewlett Packard Enterprise (HPE) open-sources the Swarm Learning Library (SLL) in binary format, the core source code of the SL framework, especially blockchain-related code, is not open-sourced. Therefore, we have to conduct black-box testing on SL and keep encapsulated features of SL untouched. For example, as mentioned in Section~\ref{Section:Implication}, we are uncertain about the underlying reasons for the unfairness of the Leader Election Algorithm (LEA) and have difficulty polishing it because it is not open-sourced. Besides, SL uses a blockchain to securely onboard members, dynamically elect the leader, and merge model parameters~\cite{warnat2021swarm}. However, these features are not open-sourced, therefore, we are unknown about the communication pattern of blockchain, the model aggregation algorithm, \textit{etc}.

\textbf{Failure of measuring connectivity.} Actually, we design an extra research question to measure the connectivity of SL. In this research question, we simulate a real-world scenario when SL nodes randomly enroll and quit the SL framework, to measure the prediction accuracy of SL. We set a connection probability $p$ for each node to individually decide whether to connect to the SL framework during each Synchronization Interval (SI). The other experimental settings are similar to those of Section~\ref{Section:RQ1}. Note that users can define \textit{min\_peers} to specify the minimum number of peers required for synchronization in the SL framework. If the online SL nodes are fewer than \textit{min\_peers}, the synchronization process would be blocked until the required number of online peers is satisfied. Therefore, as for the relationship between the number of online SL nodes ($N$) and prediction accuracy, we expect that only when $N$ is above \textit{min\_peers}, would the SL framework start to work, and the accuracy would increase as $N$ increases. Besides, we expect that only when $N$ is above a convergence threshold $N_0$, where $N_0\geq min\_peers$, would the SL framework start to converge because of the sufficiency of training data.

However, we fail to carry out this research question due to the following reasons. On the one hand, if we treat each SL node as a node once enrolled in the SL framework, when we shut down its network connection, it would only retry several times before its container is automatically shut down and removed, and the dataset and learned parameters in its container would be wasted. On the other hand, if we simply abandon the dataset and parameters in the container of each disconnected SL node and treat it as a newly enrolled one, we would hit the wall of the capacity of licenses. HPE limits the capacity of licenses assigned for non-commercial use to have at most 4 SN nodes and 16 SL nodes, and the licenses of disconnected nodes could not be revoked immediately. A new SL node is supposed to wait for at least 30 minutes for the expiry of license tokens of disconnected nodes to be assigned a new license for enrollment. The limited license capacity prevents us from continuously enrolling new nodes into the SL framework. Besides, the Synchronization Intervals (SIs) of most experiments are shorter than 30 minutes, which means SL nodes would waste redundant time in waiting for the assignment of a new license. In future work, we will try to cooperate with HPE to measure the connectivity of SL and improve the robustness of SL in face of network disconnection.

\textbf{Differences between the results of SL and FL.} As mentioned in Section~\ref{Section:Formulation}, the convergence accuracy of SL and FL is theoretically the same. However, except for the empirical study of Yang \textit{et al.}~\cite{DBLP:conf/www/YangWXCBLL21}, which focuses on the impacts of heterogeneity in \textit{FL}, there are few empirical studies comprehensively measuring the performance, imbalance, fairness, fault tolerance, and scalability of either FL or SL, as far as we are concerned. Therefore, it provides researchers and developers with valuable suggestions to measure those properties of \textit{SL}. Besides, the architectures of FL and SL are essentially different because of the introduction of blockchain, which gives us more reasons to demystify SL.


\textbf{Threats to the number of nodes.} The experiments of the first 4 research questions are conducted when the number of SN nodes is 1 and the number of SL nodes is 3 or 4. Therefore, some might argue that the results are limited to the number of nodes, which can be disassembled into two aspects, i.e. prediction accuracy and resource overhead. First, as mentioned in Section~\ref{Section:Formulation}, theoretically, when the dataset is fixed, the number of nodes would not affect convergence accuracy significantly. Second, the findings of Section~\ref{Section:RQ5} show that as the number of SN nodes increases, resource overhead of each node would remain basically unchanged, while as the number of SL nodes increases, computational overhead and network overhead would increase linearly. Therefore, as SN nodes or SL nodes scale, resource overhead could be easily inferred.

\section{Related Work} \label{Section:RelatedWork}
In this section, we introduce related work about newly emerging distributed machine learning paradigms, i.e. FL and SL, and state the position of our paper.

\textbf{Federated learning (FL).} To allow users to collectively reap the benefits of shared models trained from rich data without transmitting raw data, McMahan \textit{et al.} propose FL~\cite{DBLP:conf/aistats/McMahanMRHA17}. To improve communication efficiency, Kone{\v{c}}n{\`y} \textit{et al.}~\cite{DBLP:journals/corr/KonecnyMRR16} propose structured update and sketched update. Bonawitz \textit{et al.}~\cite{DBLP:conf/mlsys/BonawitzEGHIIKK19} introduce the protocol of FL, detailed system design on devices and servers, and some specific challenges of implementation. Yang \textit{et al.}~\cite{DBLP:journals/tist/YangLCT19} categorize FL into Horizontal FL, Vertical FL, and Federated Transfer Learning. Zhang \textit{et al.}~\cite{DBLP:conf/www/ZhangWP21} propose a FL incentive mechanism based on reputation and reverse auction theory. Liu \textit{et al.}~\cite{DBLP:conf/www/Liu0C21} adapt trained models with similar data distributions to achieve better personalization results for FL. As for applications, researchers utilize FL to help query suggestion~\cite{DBLP:journals/corr/abs-1812-02903}, keyboard prediction~\cite{DBLP:journals/corr/abs-1811-03604}, health data analytics~\cite{DBLP:conf/www/MaZ00H21}, human activity recognition~\cite{DBLP:conf/www/LiNJZY21}, and user modeling~\cite{DBLP:conf/www/Wu0HNWCYZ21}. As for empirical studies, Yang \textit{et al.}~\cite{DBLP:conf/www/YangWXCBLL21} conduct the first empirical study to characterize the impacts of heterogeneity in FL. However, FL uses central custodians to keep model parameters, which could still be attacked to infer users' identities and interests~\cite{DBLP:conf/infocom/WangSZSWQ19, DBLP:conf/sp/NasrSH19, DBLP:journals/jsac/SongWZSWRQ20, DBLP:journals/tifs/ContiMSV16, DBLP:conf/cns/WangYKH15}, even with privacy-preserving deep learning techniques, such as the shared model~\cite{DBLP:conf/iotdi/RodriguezWZMH18}, multi-party computation~\cite{goldreich1998secure}, transformation~\cite{DBLP:journals/tifs/ZhaoWZZC20, DBLP:conf/sose/FuWXMW19}, partial sharing~\cite{DBLP:conf/atis/PhongA0WM17}, model splitting~\cite{DBLP:journals/tifs/DongWWG18}, and encryption~\cite{DBLP:journals/ejisec/FontaineG07, DBLP:journals/corr/XieBFGLN14}. Besides, the star-shaped architecture of FL damages fault tolerance.

\textbf{Decentralized Federated Learning and Swarm Learning (SL).} To mitigate security and fault tolerance concerns of FL, Lalitha \textit{et al.}~\cite{lalitha2018fully, DBLP:journals/corr/abs-1901-11173} propose fully decentralized FL, where users update their belief by aggregating information from their one-hop neighbors. Ramanan \textit{et al.}~\cite{DBLP:conf/blockchain2/RamananN20} leverage Smart Contracts (SC) to coordinate the round delineation, model aggregation, and update tasks in FL. Li \textit{et al.}~\cite{DBLP:journals/corr/abs-2101-06905} allow each client to broadcast the trained model to other clients, to aggregate its own model with received ones, and then to compete to generate a block before its local training of the next round. Li \textit{et al.}~\cite{DBLP:journals/network/LiCLHZY21} use blockchain for global model storage and local model update exchange and devise an innovative committee consensus mechanism to reduce consensus computing and malicious attacks. Warnat \textit{et al.}~\cite{warnat2021swarm, warnat2020swarm} propose SL, a state-of-the-art decentralized FL paradigm that unites edge computing, blockchain-based peer-to-peer networking, and coordination while maintaining confidentiality without the need for a central coordinator, as we introduced in Section~\ref{Section:Background}. 
Cooperating with Hewlett Packard Enterprise (HPE)\footnote{https://www.hpe.com/us/en/home.html}, they develop a commercial SL software and open-source the Swarm Learning Library (SLL) in binary format for non-commercial use under evaluation license. Numerous researchers and developers are following their project on GitHub, however, they raise issues about the best practice and precautions of SL deployment\footnote{https://github.com/HewlettPackard/swarm-learning/issues}, as mentioned in Section~\ref{Section:Introduction}. 
As far as we are concerned, there are few empirical studies on SL or blockchain-based decentralized FL paradigm, therefore, we conduct this measurement to fill the knowledge gap between SL deployment and developers, and provide practical suggestions to developers and researchers. As for applications, researchers utilize SL to help diagnosis of COVID-19, tuberculosis, leukaemia and lung pathologies~\cite{warnat2021swarm},  feature selection for beta-amyloid and TAU pathology~\cite{wu2021federated}, the Internet of Vehicles~\cite{DBLP:journals/corr/abs-2108-03981}, skin lesion classification fairness~\cite{DBLP:journals/corr/abs-2109-12176}, genomics data sharing~\cite{oestreich2021privacy}, and risk prediction of cardiovascular events~\cite{westerlund2021risk}.

\section{Conclusion} \label{Section:Conclusion}
As far as we are concerned, this paper has conducted the first comprehensive study on Swarm Learning (SL), a new paradigm of blockchain-based decentralized Federated Learning (FL), to fill the knowledge gap between SL deployment and developers. We have conducted various experiments on 3 public datasets of 5 research questions, i.e. performance, imbalance, fairness, fault tolerance, and scalability. The findings have evidenced that SL is supposed to be suitable for most application scenarios, no matter whether the dataset is balanced, polluted, or biased over irrelevant features. Besides, we have quantitatively analyzed the reasons behind findings, and provided practical suggestions for developers to deploy SL on real-world applications and possible directions for researchers to optimize SL.

\clearpage
\bibliographystyle{ACM-Reference-Format}
\bibliography{ref}


\end{document}